\documentclass[runningheads]{llncs}
\usepackage{graphicx}
\usepackage{comment}
\usepackage{amsmath,amssymb} % define this before the line numbering.
\usepackage{color}

\usepackage[width=122mm,left=12mm,paperwidth=146mm,height=193mm,top=12mm,paperheight=217mm]{geometry}

\usepackage{booktabs}
\usepackage{enumitem}
\usepackage{siunitx}
\usepackage[nocompress,nospace]{cite}
\usepackage[normalem]{ulem}

\usepackage{algorithm}
\usepackage{algpseudocode}
\usepackage{makecell}

\newcommand{\ke}[1]{{\color{black}#1}}

\begin{document}
% \renewcommand\thelinenumber{\color[rgb]{0.2,0.5,0.8}\normalfont\sffamily\scriptsize\arabic{linenumber}\color[rgb]{0,0,0}}
% \renewcommand\makeLineNumber {\hss\thelinenumber\ \hspace{6mm} \rlap{\hskip\textwidth\ \hspace{6.5mm}\thelinenumber}}
% \linenumbers
\pagestyle{headings}
\mainmatter
\def\ECCVSubNumber{1932}  % Insert your submission number here

\title{Guided Collaborative Training for Pixel-wise Semi-Supervised Learning} % Replace with your title

% INITIAL SUBMISSION 
\begin{comment}
\titlerunning{ECCV-20 submission ID \ECCVSubNumber} 
\authorrunning{ECCV-20 submission ID \ECCVSubNumber} 
\author{Anonymous ECCV submission}
\institute{Paper ID \ECCVSubNumber}
\end{comment}
%******************

% CAMERA READY SUBMISSION
% \begin{comment}
\titlerunning{Guided Collaborative Training for Pixel-wise Semi-Supervised Learning}
% If the paper title is too long for the running head, you can set
% an abbreviated paper title here
%
\author{
Zhanghan Ke\inst{1,2} \and
Di Qiu\inst{2} \and
Kaican Li\inst{2} \and
Qiong Yan\inst{2} \and
Rynson W.H. Lau \inst{1}
}

\authorrunning{Z. Ke, D. Qiu, K. Li, Q. Yan and R. Lau.}
% First names are abbreviated in the running head.
% If there are more than two authors, 'et al.' is used.
%
\institute{Department of Computer Science, City University of Hong Kong \\ 
\email{kezhanghan@outlook.com, rynson.lau@cityu.edu.hk} \\ \and
SenseTime Research \\
\email{\{kezhanghan,qiudi,likaican,yanqiong\}@sensetime.com}
}
% \end{comment}
%******************
\maketitle

\setcounter{footnote}{0} 

\begin{abstract}
We investigate the generalization of semi-supervised learning (SSL) to diverse pixel-wise tasks. Although SSL methods have achieved impressive results in image classification, 
the performances of applying them to pixel-wise tasks are unsatisfactory due to their need for dense outputs. 
In addition, existing pixel-wise SSL approaches are only suitable for certain tasks as they \ke{usually} require to use task-specific properties. In this paper, we present a new SSL framework, \ke{named} Guided Collaborative Training (GCT), for pixel-wise tasks, with two main technical contributions. First, GCT addresses the issues caused by the dense outputs through a novel flaw detector. Second, the modules in GCT learn from unlabeled data collaboratively through two newly proposed constraints that are independent of task-specific properties. As a result, 
GCT can be applied to a wide range of pixel-wise tasks without structural adaptation. Our extensive experiments on four challenging vision tasks, including semantic segmentation, real image denoising, portrait image matting, and night image enhancement, show that GCT outperforms state-of-the-art SSL methods by a large margin.
Our code available at: \textcolor{blue}{\url{https://github.com/ZHKKKe/PixelSSL}}\footnote{We implemented PixelSSL, a semi-supervised learning codebase for pixel-wise tasks.}.

\end{abstract}

\begin{figure}[t]
\begin{center}
   \includegraphics[width=0.99\linewidth]{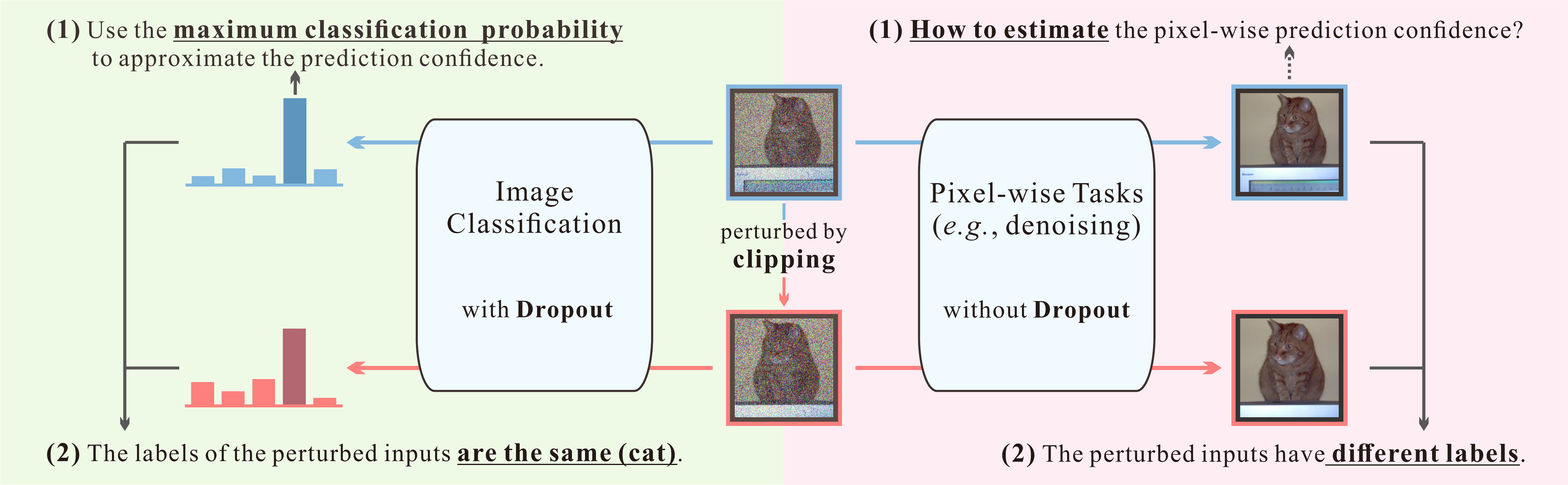}
\end{center}
\vspace{-0.5cm}
   \caption{\textbf{Difficulties of Pixel-wise SSL.} 
   The dense outputs in pixel-wise tasks causes unsatisfactory SSL performances since (1) it is difficult to estimate the pixel-wise prediction confidence and (2) existing perturbations designed for SSL are not suitable for dense outputs.
 }
\label{fig:ssl_problems}
\end{figure}

\section{Introduction}\label{sec:intro}
Deep learning has been remarkably successful in many vision tasks. Nonetheless, collecting a large amount of labeled data for training is costly, especially for pixel-wise tasks that require a precise label for each pixel, {\it e.g.}, the category mask in semantic segmentation and the clean picture in image denoising. Recently, semi-supervised learning (SSL) has become an important research direction to alleviate the lack of labels, by appending unlabeled data for training. Many SSL methods have been proposed for image classification with impressive results, including adversarial-based methods \cite{CatGAN,TripleGAN,GoodGANBadG,AdaSSL}, consistent-based methods \cite{LadderNetwork,Temporal_Pi,MeanTeacher,DualStudent}, and methods that are combined with self-supervised learning \cite{S5L, S4L}. In contrast, only a few works have applied SSL to specific pixel-wise tasks \cite{AdvSemiSeg,HighLowSemiSeg,SemiFaceSketch, UniversalSSLSeg}, and they mainly focus on semantic segmentation.

In this work, we investigate the generalization of SSL to diverse pixel-wise tasks.
Such generalization is important in order for SSL to be used in new vision tasks with minimal efforts. However, generalizing existing pixel-wise SSL methods is not straightforward since they are designed for certain tasks by using task-specific properties (Sec.\,\ref{sec:2_2}), {\it e.g.}, assuming similar semantic contents between the input and output. Another possible generalization approach is to apply SSL methods designed for image classification to pixel-wise tasks. But there are two critical issues caused by the dense outputs, as illustrated in Fig.\,\ref{fig:ssl_problems}, leading to unsatisfactory performances of these methods on pixel-wise tasks.

First, dense outputs require pixel-wise prediction confidences  (Sec.\,\ref{sec:prediction_confidence}), which are difficult to estimate. Pixel-wise tasks are either pixel-wise classification ({\it e.g.}, semantic segmentation and shadow detection) or pixel-wise regression ({\it e.g.}, image denoising and matting). Although we may use the maximum classification probability to represent the prediction confidence in pixel-wise classification, it is unavailable in pixel-wise regression.
Second, existing perturbations designed for SSL (Sec.\,\ref{sec:perturbations}) 
are not suitable for dense outputs. In pixel-wise tasks, 
strong perturbations in the input, {\it e.g.}, clipping in Mean Teacher \cite{MeanTeacher}, will change the input image and its labels. As a result, the perturbed inputs from the same original image have different labels, which is undesirable in SSL. Besides, the perturbations through Dropout \cite{Dropout} are disabled in most pixel-wise tasks. \ke{Although} Dual Student \cite{DualStudent} proposes to create perturbations through different model initializations, 
its training strategy can only be used in image classification.

To address the above two issues caused by dense outputs, 
we propose a new SSL framework, named Guided Collaborative Training (GCT), for pixel-wise tasks. It includes three modules -- two models for the specific task (the task models) and a novel flaw detector. GCT overcomes the two issues by: (1) approximating the pixel-wise prediction confidence by the output of the flaw detector, {\it i.e.}, a flaw probability map, and (2) extending the perturbations used in Dual Student to pixel-wise tasks.
Since different model initializations lead to inconsistent predictions for the same input, we can ensemble the reliable pixels, {\it i.e.}, the pixels with lower flaw probabilities, in the predictions.
In addition, minimizing the flaw probability map should help correct the unreliable pixels in the predictions.
Motivated by these ideas, we introduce two SSL constraints, a dynamic consistency constraint between the task models and a flaw correction constraint between the flaw detector and each of the task models, to allow the modules in GCT to learn from unlabeled data collaboratively under the guidance of the flaw probability map rather than the task-specific properties. As a result, GCT can be applied to diverse pixel-wise tasks, simply by replacing the task models without structural adaptations.

We evaluate GCT on the standard benchmarks for semantic segmentation (pixel-wise classification) and real image denoising (pixel-wise regression). We also conduct experiments on our own practical datasets,
{\it i.e.}, the datasets with a large proportion of unlabeled data,
for portrait image matting and night image enhancement (both are pixel-wise regression) to demonstrate the generalization of GCT on real applications. GCT surpasses start-of-the-art SSL methods \cite{MeanTeacher, AdvSemiSeg, S4L} that can be applied to these four challenging pixel-wise tasks. 
We envision that this work will contribute to future research and development of new vision tasks with scarce labels.

\section{Related Work}

\subsection{SSL for Image Classification}\label{sec:2_1}
% \textbf{SSL_i for Image Classificaion.} 
Our work is related to two main branches of SSL methods designed for image classification. The adversarial-based methods \cite{CatGAN,TripleGAN,GoodGANBadG,AdaSSL} assemble the discriminator from GAN \cite{GAN}, and try to match the latent distributions between labeled and unlabeled data through the image-level adversarial constraint. The consistent-based methods  \cite{LadderNetwork,Temporal_Pi,MeanTeacher,DualStudent} learn from  unlabeled data by applying a consistency constraint to the predictions under different perturbations. Apart from them, some latest works combine self-supervised learning with SSL \cite{S5L, S4L} or expand the training set by interpolating labeled and unlabeled data \cite{MixMatch,ReMixMatch}.

\subsection{SSL for Pixel-wise Tasks}\label{sec:2_2}
\ke{Existing} research on pixel-wise SSL mainly focuses on semantic segmentation. GANs dominate in this topic through the combination with the SSL methods derived from image classification. For example, Hung {\it et al.} \cite{AdvSemiSeg} extract reliable predictions to generate pseudo labels for training. Mittal {\it et al.} \cite{HighLowSemiSeg} modify Mean Teacher \cite{MeanTeacher} to a multi-label classifier and use it as a filter to remove uncertain categories. 
Besides, Lee {\it et al.} \cite{DSRG} and Huang {\it et al.} \cite{FickleNet} study weak-supervised learning in the SSL context. However, these works require pre-defined categories, which is a general property of classification-based tasks.
Chen {\it et al.}  \cite{SemiFaceSketch} apply SSL in face sketch synthesis, which belongs to pixel-wise regression. It regards the pre-trained VGG \cite{VGG} network as a feature extractor to impose a perceptual constraint on the unlabeled data. Unfortunately, the perceptual constraint can only be used in tasks that have similar semantic contents between the inputs and outputs. For example, it does not work on segmentation since the semantic content of the category mask is different from the input image. 

\subsection{Prediction Confidence in SSL}\label{sec:prediction_confidence}
Prediction confidence is necessary for computing the SSL constraints, which consider the predictions with higher confidence values as the targets, {\it i.e.}, pseudo labels. Earlier works show that the averaged targets are more confident. For example, Temporal Model \cite{Temporal_Pi} accumulates the predictions over epochs as the targets; Mean Teacher \cite{MeanTeacher} defines an explicit model by exponential moving average to generate the targets; FastSWA \cite{FastSWA} further averages the models between epochs to produce better targets. Others \cite{CatGAN, TripleGAN, SmoothNeighbor} regard the maximum classification probability as the prediction confidence. 

In pixel-wise SSL, the outputs of the discriminator are used to approximate the prediction confidence \cite{AdvSemiSeg,HighLowSemiSeg}. Instead, we propose the flaw detector to estimate the prediction confidence, with two key differences.
First, the flaw detector predicts a dense probability map with location information while the discriminator predicts an image-level probability. Second, we use the ground truth of the labeled data to generate the targets of the flaw detector. 

\subsection{Perturbations in SSL}\label{sec:perturbations}
Many SSL methods heavily rely on perturbations for training. The consistent-based methods \cite{Temporal_Pi,MeanTeacher,DeepCoTrain} utilize data augmentations to alter the inputs. To further improve the inconsistency, VAT \cite{VAT} generates virtual adversarial noises while S4L \cite{S4L} adds a rotation operation to the inputs. Others such as MixMatch \cite{MixMatch} and  ReMixMatch \cite{ReMixMatch} generate perturbed samples by data interpolation. Apart from the perturbations in the inputs, Dropout perturbs the predictions through a random selection of nodes \cite{VDT}. The models in Dual Student \cite{DualStudent} have inconsistent predictions for the same input due to different initializations.

\ke{Since the perturbations from both data augmentations and Dropout are not suitable for dense outputs,} GCT follows Dual Student in creating perturbations. However, unlike Dual Student, GCT learns from unlabeled data through the two SSL constraints based on the flaw detector, allowing GCT to be applicable to diverse pixel-wise tasks.

\section{Guided Collaborative Training}
\label{sec:gct}

\subsection{Overview of GCT}\label{sec:gct_overview}
\begin{figure}[t]
\begin{center}
   \includegraphics[width=0.99\linewidth]{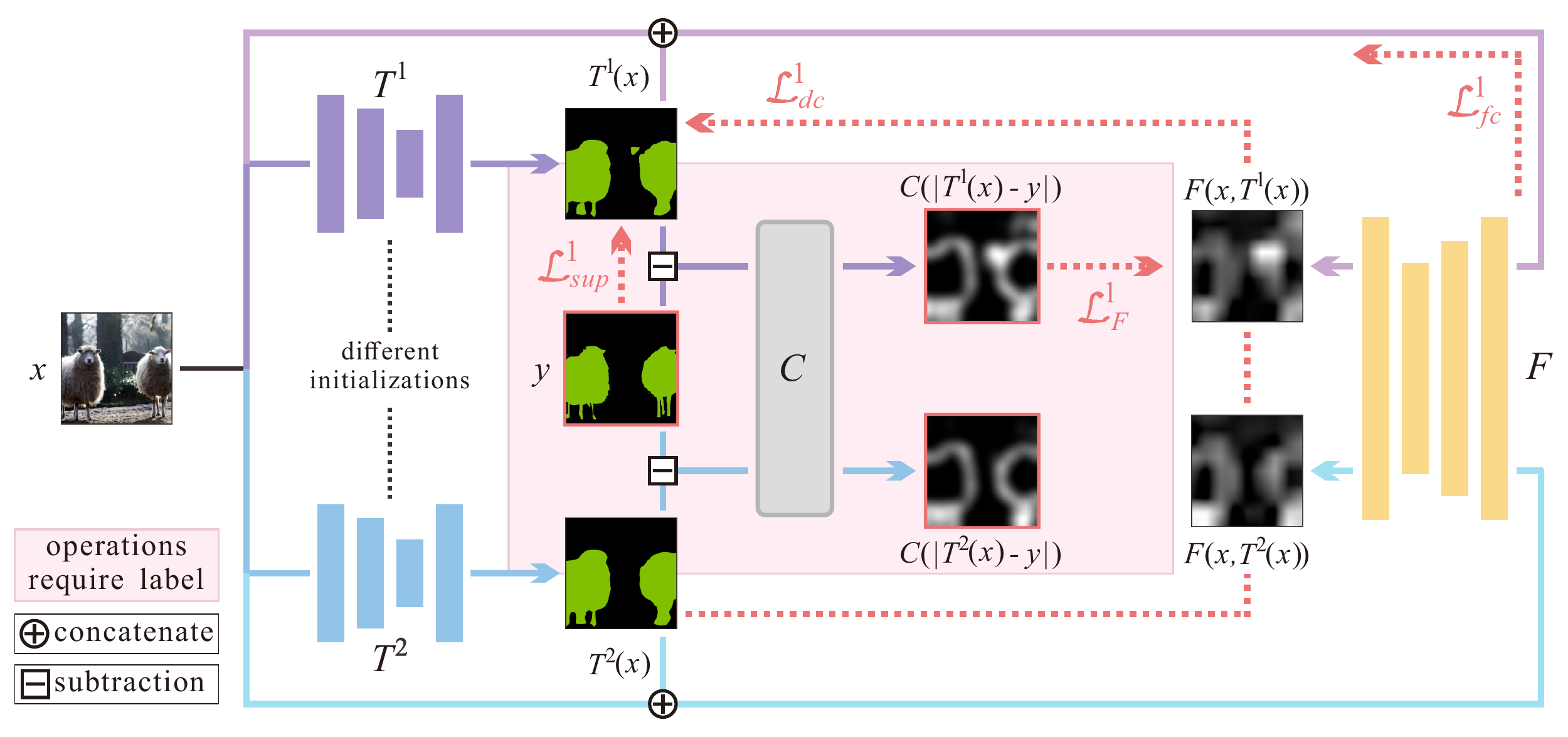}
\end{center}
\vspace{-0.5cm}
   \caption{\textbf{The GCT Framework.} It consists of two task models $T^{1}, T^{2}$ and a flaw detector $F$. Since $T^{1}$ and $T^{2}$ have different initializations, their predictions for the same $x$ are inconsistent. 
   These two task models learn from the unlabeled data through $\mathcal{L}_{dc}$ and $\mathcal{L}_{fc}$ under the guidance of the outputs of $F$. The ground truth of $F$ is calculated on the labeled subset by an image processing pipeline $C$, which takes $T^{1}(x)$ (or $T^{2}(x)$) and $y$ as the input. Here we take semantic segmentation as an example.
}
\label{fig:framework}
\end{figure}

In this section, we first present an overview of GCT. We then introduce the flaw detector and the two proposed SSL constraints.
Fig.\,\ref{fig:framework} shows the GCT framework. 
% $T^{k}$ represents either task model while $T^{2}$ represents the other. 
$T^{1}$ and $T^{2}$ are the two task models, which are referred to as $T^{k}$ ($k \in \{1, 2\}$) in the following context.
The architecture of $T^{k}$ is arbitrary, and 
% two task models with different architectures are allowed in GCT. 
GCT allows the task models to have different architectures.
The only requirement is that $T^{1}$ and $T^{2}$ should have different initializations to form the perturbations between them \ke{(which is the same as Dual Student)}. 
$F$ denotes the flaw detector. In SSL, we have a dataset consisting of a labeled subset $\mathcal{X}_{l}$ with labels $\mathcal{Y}$ and an unlabeled subset $\mathcal{X}_{u}$. The inputs $\mathcal{X} = \mathcal{X}_{l}\cup\mathcal{X}_{u}$ for both $T^{1}$ and $T^{2}$ are exactly the same. 
Given an $x \in \mathcal{X}$, the GCT framework first predicts $T^{k}(x)$ of size $H \times W \times O$, where the value of $O$ is defined by the specific task. 
Then, the concatenation of $x$ and $T^{k}(x)$ is processed by $F$ to estimate the flaw probability map $F(x, T^{k}(x))$ of size $H \times W \times 1$.
% Since the values in $F(x, T^{k}(x))$ range between $[0, 1]$, 
The prediction confidence map  
% each pixel 
can be approximated by 
% $1 - F(x, T^{k}(x))^{(h,w)}$.
$1 - F(x, T^{k}(x))$.
% Then, the flaw probability map $F(x, T^{k}(x))$ is estimated by inputting a concatenation of $x$ and $T^{k}(x)$ into $F$. 
% $F(x, T^{k}(x))$ is a $H \times W \times 1$ map that activates the flaw regions of $T^{k}(x)$.}
% To train GCT, two steps are iterated like GAN \cite{GAN}.
We train GCT iteratively in two steps like GAN \cite{GAN}. 
% Since the operations of $T^{k}$ and $T^{2}$ are symmetric, following we use $T^{k}$ as an example to explain GCT.

In the first step, we train $T^{k}$ with fixed $F$. 
% We assume that the output shape of $T^{k}$ is $H \times W \times C$.
For the labeled data, the prediction $T^{k}(x_{l})$ is supervised by its corresponding label $y$ as: 
\begin{equation}\label{eq:L_task}
  \mathcal{L}^{k}_{sup}(x_{l}, y) =  \sum_{h, w, o} \mathcal{R} (T^{k}(x_{l})^{(h,w,o)},\;y^{(h,w,o)}),
\end{equation}
where $\mathcal{R}(\cdot, \cdot)$ is a task-specific constraint, and $(h,w,o)$ is a pixel index. To learn the unlabeled data, we propose a dynamic consistency constraint $\mathcal{L}_{dc}$ and a flaw correction constraint $\mathcal{L}_{fc}$, which are guided by the flaw probability map and will be described in Sec.\,\ref{sec:L_dc} and Sec.\,\ref{sec:L_fc}, respectively. The final constraint for $T^{k}$ is a combination of three constraints as: 
\begin{equation}\label{eq:L_t}
  \mathcal{L}^{k}_{T} (\mathcal{X}, \mathcal{Y}) = \sum_{\{x_{l},y\}} \mathcal{L}^{k}_{sup} (x_{l}, y) + \sum_{x} \Big(\lambda_{dc}\,\mathcal{L}^{k}_{dc} (x) + \lambda_{fc}\,\mathcal{L}^{k}_{fc} (x)\Big),
\end{equation}
where $\{x_{l}, y\}$ is a pair of labeled data. $\lambda_{dc}$ and $\lambda_{fc}$ are hyper-parameters to balance the two SSL constraints.

In the second step, $F$ learns from the labeled subset. 
% It concatenates $x_{l}$ and $T^{k}(x_{l})$ as input and predicts a flaw probability map with shape $H \times W \times 1$.
% for $T^{k}(x_{l})$. 
We calculate the ground truth of $F$ through a classical image processing pipeline $C$ based on $T^{k}(x_{l})$ and $y$.
% We use $T^{k}(x_{l})$ and $y$ to calculate the ground truth of $F$ by a classical image processing pipeline $C$. 
In our framework, $F$ is trained by using Mean Square Error (MSE) as:
\begin{equation}\label{eq:L_d}
  \mathcal{L}^{k}_{F} (\mathcal{X}_{l}, \mathcal{Y}) =  \sum_{\{x_{l},y\}} \Big( \;\frac{1}{2} \sum_{h,w} \Big(F(x_{l}, T^{k}(x_{l}))^{(h,w)} - C(\,|T^{k}(x_{l}) - y|\,)^{(h,w)}\Big)^{2}\;\Big),
\end{equation}
where $C(|T^{k}(x_{l}) - y|)$ is the ground truth of $F$, which will be discussed in Sec.\,\ref{sec:flaw_detector}.

\begin{figure}[t]
\begin{center}
   \includegraphics[width=0.99\linewidth]{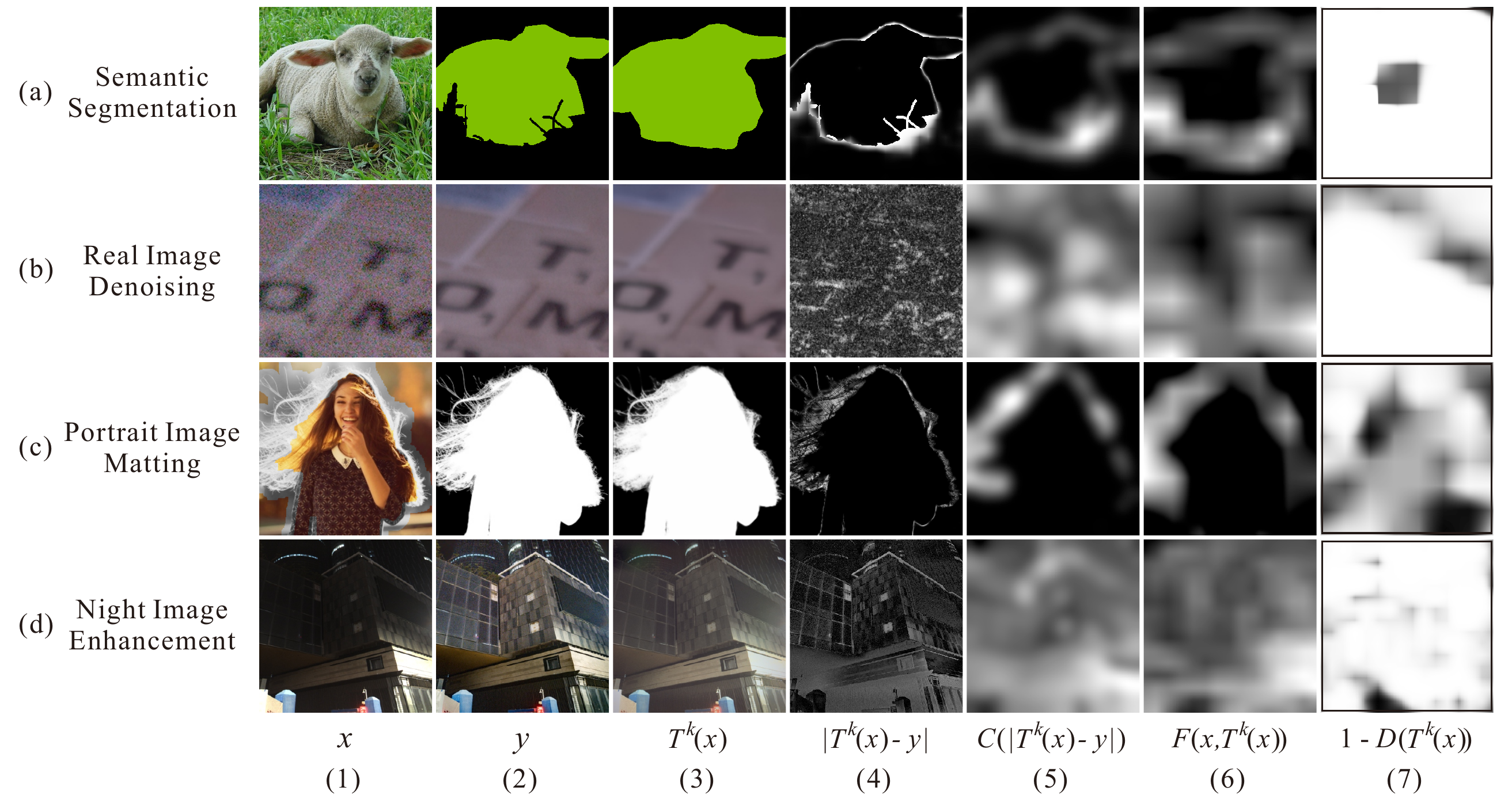}
\end{center}
\vspace{-0.5cm}
   \caption{\textbf{Flaw Detector {\it vs.} Discriminator.} 
   The flaw detector $F$ outputs $F(x, T^{k}(x))$ that highlights the flaw regions of $T^{k}(x)$ correctly. However, the fully convolutional discriminator $D$ tends to activate all small errors. We show $1-D(T^{k}(x))$ that activates the fake probability of each pixel, which has a similar meaning to the flaw probability.
   Since $|T^{k}(x) - y|$ is sparse and sharp, we use $C(|T^{k}(x) - y|)$ as the ground truth of $F$.}
    
\label{fig:flaw_map}
\end{figure}

\subsection{Flaw Detector}
\label{sec:flaw_detector}
On the labeled subset, the goal of the flaw detector $F$ is to learn the flaw probability map $F(x_{l}, T^{k}(x_{l}))$  that indicates the difference between $T^{k}(x_{l})$ and $y$, {\it i.e.}, the flaw regions in $T^{k}(x_{l})$. One simple way to find the flaw regions is $|T^{k}(x_{l}) - y|$. However, it is difficult to learn many tasks since it is sparse and sharp \ke{(column (4) of Fig.\,\ref{fig:flaw_map})}. To address this problem, we introduce an image processing pipeline $C$ that converts $|T^{k}(x_{l}) - y|$ to a dense probability map (column (5) of Fig.\,\ref{fig:flaw_map}).
$C$ consists of three basic image processing operations: dilation, blurring and normalization\footnote{Refer to Appendix\,A in the Supplementary for the algorithm of $C$.}.  
To estimate the flaw probability map $F(x_{u}, T^{k}(x_{u}))$ for the unlabeled data, we apply a common SSL assumption \cite{SSL_Survey}: the distribution of unlabeled data is the same as that of the labeled data. Therefore, $F$ trained on the labeled subset should also work well on the unlabeled subset.

% The flaw detector has a similar architecture to the discriminator $D$ in \cite{AdvSemiSeg}. 
The architecture
\footnote{Refer to Appendix\,B in the Supplementary for the architecture of the flaw detector.} 
of the flaw detector is similar to the fully convolutional discriminator $D$ in \cite{AdvSemiSeg}. 
However, $D$ averages all predicted pixels to get a single confidence value during training, as its target is an image-level real or fake probability.
% However, $D$ sums up all predicted pixels to get an average prediction confidence for training because it is trained by image-level real or fake targets.
In pixel-wise tasks, the prediction is usually accurate for some pixels but not the others, and pixels of higher accuracy should have higher confidence.
Using an average confidence to represent the overall confidence is not appropriate. For example, $T^{1}(x)$ may be more confident (more accurate) than $T^{2}(x)$ in a small local region although the average prediction confidence of $T^{1}(x)$ is lower than $T^{2}(x)$. Therefore, the per-pixel prediction confidence (from the flaw detector) is more meaningful than the average prediction confidence (from the discriminator) in pixel-wise tasks. 
Fig.\,\ref{fig:flaw_map} visualizes the results of $F$ and $D$ in the four validated tasks.

\subsection{Dynamic Consistency Constraint}
\label{sec:L_dc}

\begin{figure}[t]
\begin{center}
  \includegraphics[width=0.99\linewidth]{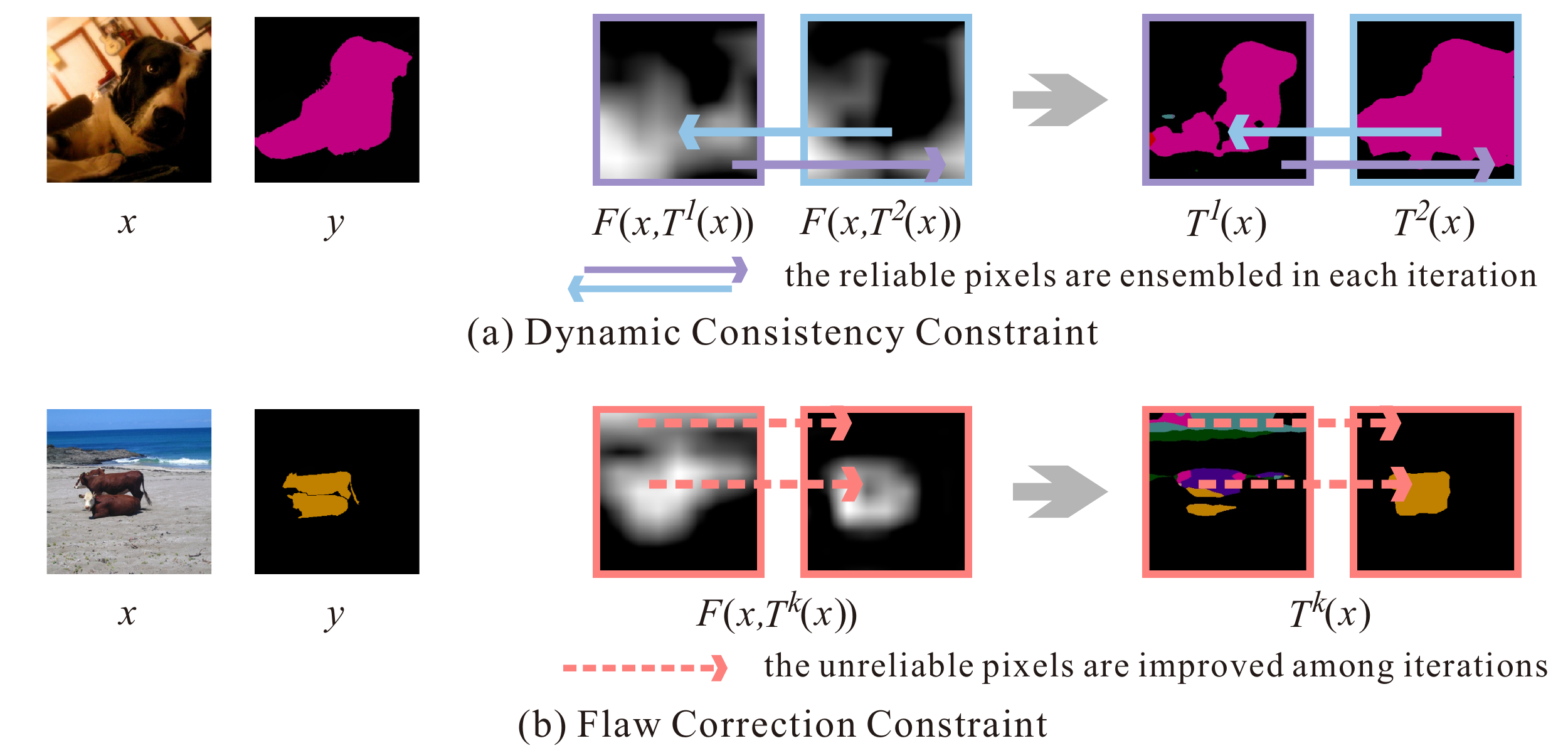}
\end{center}
\vspace{-0.5cm}
  \caption{\textbf{Proposed SSL Constraints.} (a) The dynamic consistency constraint exchanges the confident knowledge between the task models. (b) The flaw correction constraint minimizes the flaw probability map for each task model.}
\label{fig:ssl_constraints}
\end{figure}

The two task models in GCT have inconsistent predictions for the same input $x$ due to the perturbations between them. We use the dynamic consistency constraint $\mathcal{L}_{dc}$ to ensemble the reliable pixels in $T^{1}(x)$ and $T^{2}(x)$. Typically, the standard consistency constraint \cite{Temporal_Pi,MeanTeacher} is unidirectional, {\it e.g.}, from the ensemble model to the temporary model. Here, ``dynamic'' indicates that our $\mathcal{L}_{dc}$ is bidirectional and its direction changes with the flaw probability (Fig.\,\ref{fig:ssl_constraints}(a)). Intuitively, if a pixel in $T^{1}(x)$ has a lower flaw probability, we treat it as the pseudo label to the corresponding pixel in $T^{2}(x)$. To assure the quality of the pseudo label, we introduce a flaw threshold $\xi \in [0, 1]$ to disable $\mathcal{L}_{dc}$ for the pixels that have higher flaw probability values than $\xi$ in both $T^{1}(x)$ and $T^{2}(x)$. 
Through this process, there is an effective knowledge exchange between the task models,
% In this process, effective knowledge exchanging exists between the task models, 
making them collaborators.

Formally, given a sample $x \in \mathcal{X}$, GCT outputs $T^{1}(x)$, $T^{2}(x)$, and their corresponding flaw probability maps $F(x, T^{1}(x))$, $F(x, T^{2}(x))$ through forward propagation. We first normalize the values in $F(x, T^{k}(x))$ to $[0, 1]$, and then set the pixels that are larger than $\xi$ to $1$ as:
\begin{equation}
F(x, T^{k}(x))^{(h,w)} \gets \max{(F(x, T^{k}(x))^{(h,w)}, \; \big\{F(x, T^{k}(x))^{(h,w)} > \xi \big\}_{1})}.
\end{equation}
$\{condition\}_{1}$ is a boolean-to-integer function, which outputs 1 when the $condition$ is true and 0 otherwise. 
We define the dynamic consistency constraint for $T^{k}$ as:
\begin{equation}\label{eq:L_dc}
\begin{split}
       & \mathcal{L}^{k}_{dc} (x) = \frac{1}{2} \sum_{h,w} \Big(\,m^{k}_{dc}(x)^{(h,w)}\,\sum_{o} \Big(T^{k}(x)^{(h,w,o)} - T^{\tilde{k}}(x)^{(h,w,o)}\Big)^{2}\,\Big), \\
& \text{where}\qquad\; m^{k}_{dc}(x)^{(h,w)} = \big\{F(x, T^{k}(x))^{(h,w)} > F(x, T^{\tilde{k}}(x))^{(h,w)}\big\}_{1}. 
\end{split}
\end{equation}
$\tilde{k}$ represents the other task model, {\it e.g.}, $\tilde{k} = 2$ when $k = 1$.
% where $\Sigma_{\xi}$ is a matrix of the same size as $F(x, T^{k}(x))$, and the values in $\Sigma_{\xi}$ are all equal to $\xi$. 
If a flaw probability value in $F(x, T^{\tilde{k}}(x))$ is smaller than both $\xi$ and the corresponding pixel in $F(x, T^{k}(x))$, $T^{k}$ will learn this pixel from $T^{\tilde{k}}$ through $\mathcal{L}^{k}_{dc}$. We use MSE since it is widely used in SSL and is general enough for many tasks. To prevent unreliable knowledge exchange at the beginning of training, we apply a cosine ramp-up operation with $\eta$ epochs (from the standard consistency constraint) to $\mathcal{L}_{dc}$.

\subsection{Flaw Correction Constraint}
\label{sec:L_fc}

Apart from $\mathcal{L}_{dc}$, the flaw correction constraint $\mathcal{L}_{fc}$ attempts to correct the unreliable predictions of the task models (Fig.\,\ref{fig:ssl_constraints}(b)).
% we also introduce the flaw correction constraint $\mathcal{L}_{fc}$ between each of the two task models and the flaw detector (Fig.\,\ref{fig:ssl_constraints}(b)). 
The key idea behinds $\mathcal{L}_{fc}$ is to force the values in the flaw probability map to become zero. We define $\mathcal{L}_{fc}$ for $T^{k}$ (with $F$ being fixed) as:
\begin{equation}\label{eq:L_fc}
  \mathcal{L}^{k}_{fc}(x) = \frac{1}{2} \sum_{h,w} \Big( m_{fc}(x)^{(h,w)} \Big(\,F(x, T^{k}(x))^{(h,w)} - 0\,\Big) ^ {2} \Big).
\end{equation}
% In practice, w
We use a binary mask $m_{fc}(x)$ to enable $\mathcal{L}_{fc}$ on the pixels without $\mathcal{L}_{dc}$, \ke{{\it i.e.}, the pixels with unreliable predictions in both task models}:
\begin{equation}
    m_{fc}(x)^{(h,w)} = \big\{F(x, T^{1}(x))^{(h,w)} > \xi \;\;\text{AND}\;\; F(x, T^{2}(x))^{(h,w)} > \xi\big\}_{1}.
\end{equation}
% {\it i.e.}, the pixels without $\mathcal{L}_{dc}$.
% This mask prevents performance degradation caused by the overlapping of the two SSL constraints in some tasks.

We consider that the flaw detector $F$ helps improve the task models through $\mathcal{L}_{fc}$. For a system containing only one task model and the flaw detector, the objectives 
% on the labeled subset 
can be derived from Eq.\,(\ref{eq:L_d}) and (\ref{eq:L_fc}) as:
\begin{equation}\label{eq:GCT_objective}
\begin{split}
        & \min_{F}\,V_{GCT}(F) = \frac{1}{2} \mathbb{E}_{\{x_{l}, y\}\thicksim P_{\mathcal{X}_{l}, \mathcal{Y}}}\,[(F(x_{l}, T^{k}(x_{l})) - C(\,|T^{k}(x_{l}) - y|\,))^{2}], \\
        & \min_{T}\,V_{GCT}(T^{k}) = \frac{1}{2} \mathbb{E}_{x\thicksim P_{\mathcal{X}}}\,[(F(x, T^{k}(x)) - 0)^{2}], \\
\end{split}
\end{equation}
% where $\mathbb{E}$ is expectation. 
where $\mathcal{X}_{l}$ and $\mathcal{X}$ have the same distribution.
We simplify Eq.\,(\ref{eq:GCT_objective}) by removing the pixel summation operation. 
In such situation, $F$ learns the flaw probability map while $T^{k}$ optimizes it with a zero label. 
If we assume that the training process converges to an optimal solution in iteration $t^{*}$, we have: 
\begin{equation}
  \lim\nolimits_{t\to t^{*}} C(|T^{k}(x_{l}) - y|) = 0 \quad \Rightarrow \quad \lim\nolimits_{t\to t^{*}} V_{GCT}(F) = V_{GCT}(T^{k}),
\end{equation}
where $t$ is the current iteration. Hence, the objective $V_{GCT}(F)$ changes during the training process and is equal to $V_{GCT}(T^{k})$ when $t=t^{*}$. The alignment in the objectives indicates that $F$ and $T^{k}$ are collaborative to some degree.

To illustrate the difference between $\mathcal{L}_{fc}$ and the adversarial constraint, we compare Eq.\,(\ref{eq:GCT_objective}) with the objectives of LSGAN \cite{LSGAN}. If we modify LSGAN for SSL, its objectives should be:
\begin{equation}
\begin{split}
        & \min_{D}\,V_{LSGAN}(D) = \frac{1}{2} \mathbb{E}_{x\thicksim P_{\mathcal{X}}}\,[(D(T^{k}(x)) - 1)^{2}] + \frac{1}{2} \mathbb{E}_{y\thicksim P_{\mathcal{Y}}}\,[(D(y)) - 0)^{2}], \\
        & \min_{T^{k}}\,V_{LSGAN}(T^{k}) = \frac{1}{2} \mathbb{E}_{x\thicksim P_{\mathcal{X}}}\,[(D(T^{k}(x)) - 0)^{2}],
\end{split}
\end{equation}
where $D$ is the standard discriminator that tries to differentiate $T^{k}(x)$ and $y$. In contrast, $T^{k}$ tries to match the distributions between $T^{k}(x)$ and $y$. Here we reverse the labels, {\it i.e.}, 1 for fake and 0 for real, to be consistent with Eq.\,(\ref{eq:GCT_objective}). Since the targets of $D$ are constants, we have:  
\begin{equation}
\lim\nolimits_{t\to t^{*}} V_{LSGAN}(D) \neq V_{LSGAN}(T^{k}),
\end{equation}
which means that $D$ and $T^{k}$ are adversarial during the whole training process.

\section{Experiments}
In order to evaluate our framework under different ratios of the labeled data, we experiment on the standard benchmarks for semantic segmentation and real image denoising. We also experiment on the practical datasets created for portrait image matting and night image enhancement to demonstrate the generalization of GCT in real applications. We further conduct ablation experiments to analyze various aspects of GCT.

~\\
\textbf{Implementation Details.} 
We compare GCT with the model trained by the labeled data only (SupOnly) and several state-of-the-art SSL methods that can be applied to various pixel-wise tasks: (1) the adversarial-based method proposed in \cite{AdvSemiSeg} (AdvSSL); (2) the consistent-based Mean Teacher (MT) \cite{MeanTeacher}; (3) the self-supervised SSL (S4L) \cite{S4L}. 
For AdvSSL, we remove the constraint that requires classification probability to make it compatible with pixel-wise regression. For MT, we use MSE for the consistency constraint. We do not add Gaussian noise as extra perturbations since it will degrade the performance. For S4L, a four-category classifier trained by Cross Entropy is added to the end of the task model to predict the rotation angles. 
(0\si{\degree}, 90\si{\degree}, 180\si{\degree}, 270\si{\degree}).

~\\
\textbf{Experimental Setup.} We notice that existing works of pixel-wise SSL usually report a fully supervised baseline with a lower performance than the original paper due to inconsistent hyper-parameters. In image classification, a similar situation has been discussed by \cite{SSLEval}. To fairly evaluate the performance of SSL, we define some training rules to improve the SupOnly baselines. We denote the total number of trained samples as $N = S * T * b$, where $S$ is the training epochs, $T$ is the number of iterations in each epoch, and $b$ is the batch size, which is fixed in each task. 
For the experiments performed on the standard benchmarks:
\begin{enumerate}[label=(\arabic*)]
    \item We train the fully supervised baseline according to the hyper-parameters from the original paper to achieve a comparable result. The same hyper-parameters (except $S$) are used in (2) and (3). 
    \item We use the same $S$ as in (1) to train the models supervised by the labeled subset (SupOnly). Although $T$ decreases as the labeled data reduces, to prevent overfitting, we do not increase $N$ by training more epochs.
    \item We adjust $S$ to ensure that $N$ in SSL experiments is the same as (1). In SSL experiments, each batch contains both labeled and unlabeled data. We define ``epoch'' as going through the unlabeled subset for once. Meanwhile, the labeled subset is repeated several times inside an epoch.
\end{enumerate}
By following these rules, the SupOnly baselines obtain good enough performance and do not overfit. The models trained by SSL methods have the same computational overhead, {\it i.e.}, the same $N$, as the fully supervised baseline. 
For experiments on the practical datasets, we first train $S$ epochs for the SupOnly baselines. Afterwards, we train the SSL models with the same $S$.
We use the grid search to find suitable hyper-parameters for all SSL methods.
\footnote{Refer to Appendix C in the Supplementary for more training details.}\footnote{Refer to Appendix D in the Supplementary for visual comparisons.}.

% More training details are provided in Appendix\,C in the Supplementary. 
% Unless noted, all reported results are averaged over 3 runs, and the same labeled data split is used for all SSL experiments.

\subsection{Semantic Segmentation Experiments}
Semantic segmentation \cite{FCNSS,deeplabv2,deeplabv3plus} takes an image as input and predicts a series of category masks, which link each pixel in the input image to a class (Fig.\,\ref{fig:flaw_map}(a)). 
% We merge the masks by different colors for visualizing in Fig.\,\ref{}. 
We conduct experiments on the Pascal VOC 2012 dataset \cite{pascalvoc}, which comprises 20 foreground classes along with 1 background class. The extra annotation set from the Segmentation Boundaries Dataset (SBD) \cite{SBD} is combined to expand the dataset. Therefore, we have 10,582 training samples and 1,449 validation samples. During training, the input images are cropped to $321\times321$ after random scaling and horizontal flipping. Following previous works \cite{AdvSemiSeg,HighLowSemiSeg}, we use DeepLab-v2 \cite{deeplabv2} with the ResNet-101 \cite{ResNet} backbone as the SupOnly baselines and as the task model in SSL methods. The same configurations as the original paper of DeepLab-v2 are applied, except the multi-scale fusion trick.

For SSL, we randomly extract $1/16$, $1/8$, $1/4$, $1/2$ samples as the labeled subset, and use the rest of the training set as the unlabeled subset. Note that the same data splits are used in all SSL methods. Table \ref{tab:sseg} shows the mean Intersection-over-Union (mIOU) on the PASCAL VOC 2012 dataset with pre-training on the Microsoft COCO dataset \cite{COCO}. GCT achieves a performance increase of $1.26\%$ (under $1/2$ labels) to $3.76\%$ (under $1/8$ labels) over the SupOnly baselines. Moreover, our fully supervised baseline ($75.32\%$) is comparable with the original paper of DeepLab-v2 ($75.14\%$), which is better than the result reported in \cite{AdvSemiSeg} ($73.6\%$). Therefore, all SSL methods only have slight improvement under the full labels.
% The improvements of AdvSSL (from $0.84\%$ to $1.51\%$) are lower than its original paper. The results also show the inability of S4L in semantic segmentation.

\begin{table*}[t]
  \begin{center}
    \caption{\textbf{Results of Semantic Segmentation.} We report mIOU ($\%$) on the validation set of Pascal VOC 2012 averaged over 3 runs. The task model is DeepLab-v2.}
    \label{tab:sseg}
    \begin{tabular}{lccccc}
      \toprule 
      \textbf{\;Methods\;\;\;\;\;\;\;\;\;\;} & \;\;1/16 labels\;\; & \;\;1/8 labels\;\; & \;\;1/4 labels\;\; & \;\;1/2 labels\;\; & \;\;full labels\; \\
      \midrule
        % & \multicolumn{5}{c}{Results (mIOU) of DeepLab-v2} \\
    %   \midrule
      \;SupOnly & $64.55$ & $68.38$ & $70.69$ & $73.56$ & $75.32$ \\
      \;MT \cite{MeanTeacher} & $66.08$ & $69.81$ & $71.28$ & $73.23$ & $75.28$ \\
      \;S4L \cite{S4L} & $64.71$ & $68.65$ & $70.97$ & $73.43$ & $75.38$ \\
      \;AdvSSL \cite{AdvSemiSeg} & $65.67$ & $69.89$ & $71.53$ & $74.48$ & $\mathbf{75.86}$ \\
      \;GCT (Our) & $\mathbf{67.19}$ & $\mathbf{72.14}$ & $\mathbf{73.62}$ & $\mathbf{74.82}$ & $75.73$ \\
      \bottomrule
    \end{tabular}
  \end{center}
  \vspace{-0.2cm}
\end{table*}

\begin{table*}[t]
  \begin{center}
    \caption{\textbf{Results of Real Image Denoising.} We report PSNR (dB) on the validation set of SIDD averaged over 3 runs. The task model is DHDN. }
    \label{tab:denoising}
    \begin{tabular}{lccccc}
      \toprule 
      \textbf{\;Methods\;\;\;\;\;\;\;\;\;\;} & \;\;1/16 labels\;\; & \;\;1/8 labels\;\; & \;\;1/4 labels\;\; & \;\;1/2 labels\;\; & \;\;full labels\; \\
      \midrule
      \;SupOnly & $37.52$ & $38.16$ & $38.74$ & $39.14$ & $39.38$ \\
      \;MT \cite{MeanTeacher} & $37.73$ & $38.22$ & $38.64$ & $39.08$ & $39.43$ \\
      \;S4L \cite{S4L} & $37.81$ & $38.32$ & $38.88$ & $39.21$ & $39.16$ \\ 
      \;AdvSSL \cite{AdvSemiSeg} & $37.85$ & $38.28$ & $38.83$ & $39.18$ & $39.47$ \\
      \;GCT (Our) & $\mathbf{38.13}$ & $\mathbf{38.56}$ & $\mathbf{38.96}$ & $\mathbf{39.30}$ & $\mathbf{39.51}$ \\
      \bottomrule
    \end{tabular}
  \end{center}
  \vspace{-0.2cm}
\end{table*}

\subsection{Real Image Denoising Experiments}

Real image denoising \cite{FFDNet,CBDNet,DenoisingFA} is a task that devotes to removing the real noise, rather synthetic noise, from an input natural image (Fig.\,\ref{fig:flaw_map}(b)). We conduct experiments on the SIDD dataset \cite{SIDD}, which is one of the largest benchmarks on real image denoising. It contains 160 image pairs (noisy image and clean image) for training and 40 image pairs for validation. We split each image pair into multiple patches with size $256\times256$ for training. The total training samples is about 30,000.
We use DHDN \cite{DHDN}, a method that won the second place in the NTRIE 2019 real image denoising challenge \cite{NTIRE}, as the task model since the code for the first place winner has not been published.
The peak-signal-to-noise-ratio (PSNR) is used as the validation metric.

In image denoising, even small errors between the prediction and the ground truth can result in obvious visual artifacts. It means that the reliable pseudo labels are difficult to obtain, {\it i.e.}, this task is difficult for SSL. We notice that the task models with the same architecture in GCT have similar predictions. Therefore, the perturbations from different initializations are not strong enough. To alleviate this problem, we replace one of the task models with DIDN \cite{DIDN} that won the third place in the NTRIE 2019 challenge. 
We still use DHDN for validation.  

We extract $1/16$, $1/8$, $1/4$, $1/2$ labeled image pairs randomly for SSL.
As shown in Table\,\ref{tab:denoising}, our fully supervised baseline achieves 39.38dB (PSNR), which is comparable with the top-level results on the SIDD benchmark. Although SSL shows limited performance in this difficult task, GCT surpasses other SSL methods under all labeled ratios. Notably, GCT improves on PSNR by 0.61dB with $1/16$ labels (only 10 labeled image pairs) while the previous SSL methods improve on PSNR by 0.33dB at most.

\begin{table*}[t]
  \begin{center}
    \caption{\textbf{Results of Portrait Image Matting and Night Image Enhancement.} We report PSNR (dB) on the validation set of the practical datasets averaged over 3 runs. In the table, ``L'' means labeled data while ``U'' means unlabeled data.}
    \label{tab:matting_hdr}
    \begin{tabular}{lcccc}
      \toprule 
       &  \multicolumn{2}{c}{Portrait Image Matting} & \;\;\; & Night Image Enhancement \\ \cmidrule(r){2-3} \cmidrule(r){5-5} 

      \textbf{\;Methods\;\;\;\;\;\;\;\;\;\;} & 100L + 3,850U\;\; & \;\; 100L + 7,700U\;\; & \; & 200L + 1,500U  \\
        \midrule
      \;SupOnly & $25.39$ & $25.39$ & & $18.72$  \\
      \;MT \cite{MeanTeacher} & $26.60$ & $27.63$ & & $19.93$  \\
      \;S4L \cite{S4L} & $26.87$ & $28.24$ & & $19.63$ \\
      \;AdvSSL \cite{AdvSemiSeg} & $26.52$ & $27.57$ & & $19.59$  \\
      \;GCT (Our) & $\mathbf{27.35}$ & $\mathbf{29.38}$ & & $\mathbf{20.14}$ \\
      \bottomrule
    \end{tabular}
  \end{center}
  \vspace{-0.2cm}
\end{table*}

\subsection{Portrait Image Matting Experiments}
Image Matting \cite{portraitmatting,DIM} predicts a foreground mask (matte) from an input image and a pre-defined trimap. Each pixel value in the matte is a probability between $[0, 1]$. 
We focus on the matting of portrait images here, which has important applications on smartphone, {\it e.g.}, blurring the background of an image. 
In Fig.\,3(c), the trimap is merged into $x$ for visualization by setting the pixels inside the unknown region of the trimap to gray.
Since there are no open-source benchmarks, we first collected 8,000 portrait images from Flickr.
We then generate the trimaps from the results of a pre-trained segmentation model. After that, we select 300 images with fine details and label them by Photoshop ($\sim$20min per image). Finally, we combine 100 labeled images with 7,700 unlabeled images as the training set, while the remaining 200 labeled images are used as the validation set. For each labeled image, we generate 15 samples by random cropping and 35 samples by background replacement (with the OpenImage dataset \cite{openimage}). For each unlabeled image, we generate 5 samples by random cropping. The structure of our task model is derived from \cite{DIM}, which is a milestone in image matting.

In this task, we verify the impact of increasing the amount of  unlabeled data on SSL by experimenting on two configurations. With 100 labeled images, (1) we randomly select half (3,850) of unlabeled images for training, and (2) we use all (7,700) unlabeled images for training. As shown in Table\,\ref{tab:matting_hdr}, GCT yields an improvement over the SupOnly baselines by 1.96dB and 3.99dB for 3,850 and 7,700 unlabeled images respectively. This indicates that the SSL performance can be effectively improved by increasing the amount of unlabeled data. In addition, doubling the amount of unlabeled images achieves a more significant improvement (2.03dB) with GCT, compared with existing SSL methods.

\subsection{Night Image Enhancement Experiments}
Night Image Enhancement \cite{HDRNet,LSDark} is another common vision application. This task adjusts the coefficients of the channels in a night image to show more details (Fig.\,\ref{fig:flaw_map}(d)). Our dataset contains 1,900 night images captured by smartphones, of which 400 images are labeled using Photoshop ($\sim$15min per image). We combine 200 labeled images with 1,500 unlabeled images for training and use another 200 labeled images for testing.  We use horizontal flipping, slight rotation, and random cropping (to $512\times512$) as data augmentations during training. We regard HDRNet \cite{HDRNet} as the task model. Since the dataset is small, we experimented with only one SSL configuration (Table\,\ref{tab:matting_hdr}). Similar to the experiments in the other three tasks, GCT outperforms existing SSL methods.

\begin{figure}[t]
\begin{center}
% \fbox{\rule{0pt}{2.5in} \rule{0.9\linewidth}{0pt}}
  \includegraphics[width=0.99\linewidth]{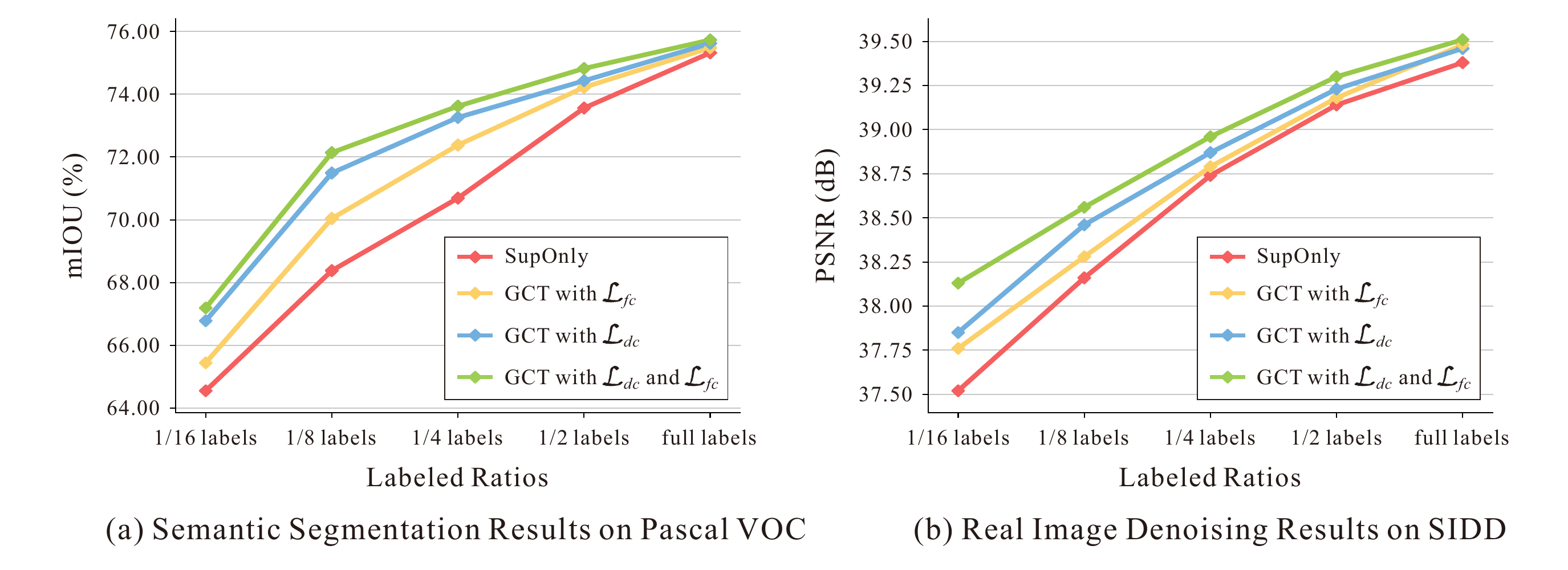}
\end{center}
\vspace{-0.5cm}
  \caption{\textbf{Ablation of the Proposed SSL Constraints.} We compare the performance of $\mathcal{L}_{dc}$ and $\mathcal{L}_{fc}$ on (a) the  Pascal VOC benchmark and (b) the SIDD benchmark. The results of SupOnly (red) and GCT with two SSL constraints (green) are the same as (a) Table\,\ref{tab:sseg} and (b) Table\,\ref{tab:denoising}.}
\label{fig:constraints_ablation}
\end{figure}

\subsection{Ablation Experiments}
\ke{We conduct} ablation studies to analyze the proposed SSL constraints, the hyper-parameters in GCT, and the combination of the flaw detector and Mean Teacher.

~\\
\textbf{Effect of the SSL Constraints.} 
By default, GCT learns from the unlabeled data through the two SSL constraints simultaneously. In Fig.\,\ref{fig:constraints_ablation}, we compare the experiments of training GCT with only one SSL constraint on the benchmarks for semantic segmentation and real image denoising. The results demonstrate that both $\mathcal{L}_{dc}$ and $\mathcal{L}_{fc}$ are effective. GCT with $\mathcal{L}_{dc}$ boosts the performance impressively, proving that the knowledge exchange between the two task models is reliable and effective. Meanwhile, the curve of GCT with $\mathcal{L}_{fc}$ indicates that the flaw detector also plays a vital role in learning the unlabeled data. Moreover, combining $\mathcal{L}_{dc}$ and $\mathcal{L}_{fc}$ allows GCT to achieve the optimal performance.

\begin{table*}[t]
  \begin{center}
    \caption{\textbf{Ablation of Hyper-Parameters.}  We report mIOU (\%) on the Pascal VOC benchmark with $1/8$ labels. The result under $\xi=0.4$ or $\eta=3$ is the same as Table\,\ref{tab:sseg}.}
    \label{tab:ablation}
 \begin{minipage}[t]{0.54\textwidth}
  \centering
    %  \makeatletter\def\@captype{table}\makeatother\caption{xx1}
       \begin{tabular}{cccccc} 
       \toprule 
       \multicolumn{6}{c}{flaw threshold $\xi$} \\
       \;\;$0.0$\;\; & \;\;$0.2$\;\; & \;\;$0.4$\;\; & \;\;$0.6$\;\; & \;\;$0.8$\;\; & \;\;$1.0$\;\; \\
       \midrule
       \;$70.04$\; & \;$70.92$\; & \;$72.14$\; & \;$\mathbf{72.43}$\; & \;$71.96$\; & \;$71.49$\; \\
       \bottomrule
    \end{tabular}
  \end{minipage}
  \begin{minipage}[t]{0.44\textwidth}
   \centering
    %  \makeatletter\def\@captype{table}\makeatother\caption{xx1}
       \begin{tabular}{ccccc} 
       \toprule 
       \multicolumn{5}{c}{ramp-up epochs $\eta$} \\
       \;\;$0$\;\; & \;\;$1$\;\; & \;\;$3$\;\; & \;\;$5$\;\; & \;\;$10$\;\; \\
       \midrule
       \;$71.34$\; & \;$72.03$\; & \;$\mathbf{72.14}$\; & \;$72.06$\; & \;$71.95$\; \\
       \bottomrule
      \end{tabular}
   \end{minipage}

  \end{center}
  \vspace{-0.2cm}
\end{table*}

~\\
\textbf{Hyper-parameters in GCT.} We analyze the two hyper-parameters required by GCT (mentioned in Sec.\,\ref{sec:L_dc}), the flaw threshold $\xi$ and the cosine ramp-up epochs $\eta$ of $\mathcal{L}_{dc}$, on the Pascal VOC benchmark for semantic segmentation with $1/8$ labels. 
Table\,\ref{tab:ablation} (left) shows the results under different $\xi$, which controls the combination of the two SSL constraints. 
Specifically, only $\mathcal{L}_{fc}$ is applied when $\xi = 0.0$, and only $\mathcal{L}_{dc}$ is applied when $\xi = 1.0$. Our experiments show that $\xi$ can be set roughly, {\it e.g.}, $\xi \in [0.4, 0.8]$ is suitable for semantic segmentation. 
The cosine ramp-up with $\eta$ epochs prevents exchanging unreliable knowledge due to the non-convergent flaw detector in the early training stage. The results in Table\,\ref{tab:ablation} (right) indicate that GCT is robust to $\eta$, even though the cosine ramp-up is necessary for the best performance.

~\\
\textbf{Combination of the Flaw Detector and MT.}
The consistency constraint in MT is applied from the teacher model to the student model. However, the teacher model may be worse than the student model on some pixels, which may cause a performance degradation. To avoid this problem, we use the flaw detector to disable the consistency constraint when the flaw probability of the teacher's prediction is larger than the student's prediction. Under $1/8$ labels, this method improves the mIOU value of MT from $69.81\%$ to $70.47\%$ on Pascal VOC and improves the PSNR value of MT from 38.22dB to 38.42dB on SIDD. 
% Our results indicate that the flaw detector has the potential to be used in other SSL methods.

\section{Conclusions}
We have studied the generalization of SSL to diverse pixel-wise tasks and indicated the drawbacks of existing SSL methods in these tasks, which to the best of our knowledge is the first. We have presented a new general framework, named GCT, for pixel-wise SSL. Our experiments have proved its effectiveness in a variety of vision tasks. Meanwhile, we also note that SSL still has limited performance for tasks that require highly precise pseudo labels, such as image denoising. A possible future work is to investigate this problem and explore ways to create more accurate pseudo labels.

\title{Guided Collaborative Training for Pixel-wise Semi-Supervised Learning \\
 ~\\
\large{Supplementary Material}} % Replace with your title

% INITIAL SUBMISSION 
\begin{comment}
\titlerunning{ECCV-20 submission ID \ECCVSubNumber} 
\authorrunning{ECCV-20 submission ID \ECCVSubNumber} 
\author{Anonymous ECCV submission}
\institute{Paper ID \ECCVSubNumber}
\end{comment}
%******************

% CAMERA READY SUBMISSION
% \begin{comment}
\titlerunning{Guided Collaborative Training for Pixel-wise Semi-Supervised Learning}
% If the paper title is too long for the running head, you can set
% an abbreviated paper title here
%
\author{
Zhanghan Ke\inst{1,2} \and
Di Qiu\inst{2} \and
Kaican Li\inst{2} \and
Qiong Yan\inst{2} \and
Rynson W.H. Lau \inst{1}
}

\authorrunning{Z. Ke, D. Qiu, K. Li, Q. Yan and R. Lau.}
% First names are abbreviated in the running head.
% If there are more than two authors, 'et al.' is used.
%
\institute{Department of Computer Science, City University of Hong Kong \\ 
\email{kezhanghan@outlook.com, rynson.lau@cityu.edu.hk} \\ \and
SenseTime Research \\
\email{\{kezhanghan,qiudi,likaican,yanqiong\}@sensetime.com}
}
% \end{comment}
%******************
\maketitle

\section*{Appendix A: Algorithm of $C$}
In GCT, we use a classical image processing pipeline $C$ to calculate the ground truth of the flaw detector $F$ on the labeled subset by taking the task model prediction $T^{k}(x_{l})$ and the corresponding label $y$ as the input. 
% $C$ is modified according to specific tasks, and it contains three main operations:
$C$ is composed of three operations:
\begin{enumerate}[itemsep=2pt]
    \item $blur(inp,\;(height, width))$: Blur $inp$ by a Gaussian kernel of given shape.
    \item $dilate(inp,\;(height, width))$: Dilate $inp$ for each local region of given shape.
    \item $norm(inp)$: Normalize all pixels in $inp$ to range between $[0,\,1]$.
\end{enumerate}
% $C$ for semantic segmentation and portrait image matting are similar (see Algorithm\;\ref{alg:C_SSEG_MATTING}) while $C$ for real image denoising and night image enhencement are similar (see Algorithm\;\ref{alg:C_DENOISING_ENHANCEMENT}).
We show the pseudo code of $C$ in Python style as follows (assume the shape of $T^{k}(x_l)$ is $H \times W \times O$):

% \vspace{-0.2cm}
\begin{algorithm}[H]
  \caption{Image Process Pipeline $C$.}
  \label{alg:C_SSEG_MATTING}
  \begin{algorithmic}[1]
  \Require Channel average coefficient $\mu$ ; Operations repeat times $\nu$. 
    \State \textbf{def} $C(T^{k}(x_{l}),\;y)$:
    \State \qquad $F_{gt} = \mu \sum_{o} |T^{k}(x_{l})^{(h, w, o)} - y^{(h, w, o)}| $
    \State \qquad $F_{gt} = blur(F_{gt}, \; (\frac{H}{8}, \frac{W}{8}))$
    \State \qquad \textbf{for} $i$ \textbf{in} $range(0,\;\nu)$:
    \State \qquad \qquad $F_{gt} = dilate(F_{gt}, \; (3, 3))$
    \State \qquad \qquad $F_{gt} = blur(F_{gt}, \; (\frac{H}{4}, \frac{W}{4}))$
    \State \qquad $F_{gt} = norm(F_{gt})$
    \State \qquad \textbf{return} $F_{gt}$
  \end{algorithmic}
\end{algorithm}
% \vspace{-0.2cm}
In our experiments, we set $\mu = \frac{1}{2}$ for semantic segmentation, and we set $\mu = \frac{1}{o}$ for other three tasks. We set $\nu = 10$ for real image denoising, $\nu = 5$ for night image enhancement, and $\nu = 1$ for other two tasks.

% \vspace{-0.2cm}
% \begin{algorithm}[H]
%   \caption{$C$ for Semantic Segmentation and Portrait Image Matting.}
%   \label{alg:C_SSEG_MATTING}
%   \begin{algorithmic}[1]
%   \Require Channel average coefficient $\mu$ ($\mu=\frac{1}{2}$ in segmentation; $\mu=1$ in matting).
%   \Require Gaussian blur coefficient $\kappa$ ($\kappa=8$ in segmentation; $\kappa=16$ in matting).
  
%     \State \textbf{def} $C(T^{k}(x_{l}),\;y)$:
%     \State \qquad $F_{gt} = \mu \sum_{o} |T^{k}(x_{l})^{(h, w, o)} - y^{(h, w, o)}| $
%     \State \qquad $F_{gt} = blur(F_{gt}, \; (\frac{H}{\kappa}, \frac{W}{\kappa}))$
%     \State \qquad $F_{gt} = dilate(F_{gt}, \; (3, 3))$
%     \State \qquad $F_{gt} = blur(F_{gt}, \; (\frac{H}{4}, \frac{W}{4}))$
%     \State \qquad $F_{gt} = norm(F_{gt})$
%     \State \qquad \textbf{return} $F_{gt}$
%   \end{algorithmic}
% \end{algorithm}
% \vspace{-1cm}

% \begin{algorithm}[H]
%   \caption{$C$ for Real Image Denoising and Night Image Enhancement.}
%   \label{alg:C_DENOISING_ENHANCEMENT}
%   \begin{algorithmic}[1]
%   \Require Operation repeat times $\nu$ ($\nu=10$ in denoising; $\nu=5$ in enhancement).
  
%     \State \textbf{def} $C(T^{k}(x_{l}),\;y)$:
%     \State \qquad $F_{gt} = \frac{1}{o} \sum_{o} |T^{k}(x_{l})^{(h, w, o)} - y^{(h, w, o)}|$
%     \State \qquad \textbf{for} $i$ \textbf{in} $range(0,\;\nu)$:
%     \State \qquad\qquad $F_{gt} = dilate(F_{gt}, \; (3, 3))$
%     \State \qquad\qquad $F_{gt} = blur(F_{gt}, \; (\frac{H}{8}, \frac{W}{8}))$
%     \State \qquad $F_{gt} = norm(F_{gt})$
%     \State \qquad \textbf{return} $F_{gt}$
%   \end{algorithmic}
% \end{algorithm}
% \vspace{-1cm}

\section*{Appendix B: Architecture of Flaw Detector}
The flaw detector $F$ is a fully-convolutional neural network, which contains 8 convolutional layers with $4 \times 4$ kernels. The amount of kernels is increased from $64$ to $512$ in the first $7$ layers and then decreased to $1$ in the last layer. Each of the first 7 convolutional layers is followed by batch normalization \cite{BatchNorm} and leaky ReLU \cite{LeakyRelU} with threshold of $0.2$.
The convolutional layers with stride=$2$ reduce the resolution of the feature maps.
At the end of $F$, we add a bilinear interpolation operation to rescale the output to the size of the input. In all experiments of GCT, we optimize $F$ by Adam \cite{Adam} (with learning rate $1e^{-4}$). The architecture of $F$ is as follow:

\begin{table}[H]
\vspace{-0.2cm}
  \begin{center}
    % \caption{\textbf{The Architecture of the Flaw Detector.} In this table, ``Conv'' indicates the convolutional layer; ``BN'' indicates the batch normalization layer; ``ReLU'' indicates the leaky ReLU layer with negative-slope=$0.2$. The shape of $T^{k}(x)$ is $H \times W \times O$.}
    \label{tab:fd_cnn}
    \begin{tabular}{ll}
      \toprule 
      \textbf{Layer} \qquad\qquad\qquad\qquad  &\textbf{Details}                                             \\
      \midrule
      Input                 & concatenate $T^{k}(x)$ and $x$ as the input                     \\ 
    %   noise                 & gaussian noise $\zeta$ = $0.15$                                     \\ 
      Conv + BN   + ReLU \;\;\;\;        & out-channels=$64$, \;\,kernel-size=$4$, stride=$2$, padding=$same$   \\
      Conv + BN   + ReLU         & out-channels=$128$, kernel-size=$4$, stride=$2$, padding=$same$   \\
      Conv + BN   + ReLU         & out-channels=$128$, kernel-size=$4$, stride=$1$, padding=$same$   \\
      Conv + BN  + ReLU          & out-channels=$256$, kernel-size=$4$, stride=$2$, padding=$same$   \\     
      Conv + BN   + ReLU         & out-channels=$256$, kernel-size=$4$, stride=$1$, padding=$same$   \\ 
      Conv + BN  + ReLU          & out-channels=$512$, kernel-size=$4$, stride=$2$, padding=$same$   \\     
      Conv + BN + ReLU         & out-channels=$512$, kernel-size=$4$, stride=$1$, padding=$same$   \\
      Conv           & out-channels=$1$, \;\;\;\,kernel-size=$4$, stride=$2$, padding=$same$   \\ 
      Interpolation         & out-shape=$H \times W$, mode=$bilinear$, align-corners=$True$ \\
      \bottomrule
    \end{tabular}
  \end{center}
  \vspace{-0.5cm}
\end{table}

\section*{Appendix C: Training Details}
We have experimented with several SSL methods, including (1) the consistent-based Mean Teacher (MT) \cite{MeanTeacher}; (2) the self-supervised SSL (S4L) \cite{S4L}; (3) the adversarial-based method proposed in \cite{AdvSemiSeg} (AdvSSL); (4) the GCT framework proposed by us. Here are the definitions of the hyper-parameters for SSL in these methods:
\begin{table}[H]
\vspace{-0.2cm}
  \begin{center}
    % \caption{\textbf{Results of Semantic Segmentation.} We report mIOU ($\%$) on the validation set of Pascal VOC 2012 averaged over 3 runs. The task model is DeepLab-v2.}
    \label{tab:ssl_hyperparameters}
    \begin{tabular}{lccl}
      \toprule 
      \textbf{\;Methods\;\;\;\;\;\;\;} & \multicolumn{3}{l}{\textbf{Hyper-Parameters for SSL}}\\
      \midrule
        % & \multicolumn{5}{c}{Results (mIOU) of DeepLab-v2} \\
    %   \midrule
      \;MT \cite{MeanTeacher} & $\lambda_{MT}$ & - & coefficient for scaling the consistency constraint \\
      & $\eta_{MT}$ &- & epochs for ramping up the consistency constraint \\
      & $\alpha_{MT}$ &- & moving average coefficient for ensembling the teacher model \\
      \midrule
      \;S4L \cite{S4L} & $\lambda_{S4L}$ &- & coefficient for scaling the unsupervised rotation constraint \\
      \midrule
      \;AdvSSL \cite{AdvSemiSeg} & $\lambda^{l}_{Adv}$ & - &coefficient for scaling the labeled adversarial constraint \\
      & $\lambda^{u}_{Adv}$ & - &coefficient for scaling the unlabeled adversarial constraint  \\
      \midrule
      \;GCT (Our) 
      & $\lambda_{fc}$ & - &coefficient for scaling the flaw correction constraint \\
      & $\lambda_{dc}$ &- & coefficient for scaling the dynamic consistency constraint \\
      & $\eta_{dc}$ &- & epochs for ramping up the dynamic consistency constraint \\
      & $\xi$ &- & flaw threshold for calculating the dynamic consistency  \\
      & &  &constraint and combining the two SSL constraints  \\
      \bottomrule
    \end{tabular}
  \end{center}
  \vspace{-0.5cm}
\end{table}
For the four validated tasks, we use grid search to find the suitable hyper-parameters for SSL. The final settings for the experiments are as follows: 
\begin{table}[H]
  \vspace{-0.2cm}
  \begin{center}
    % \caption{\textbf{Results of Semantic Segmentation.} We report mIOU ($\%$) on the validation set of Pascal VOC 2012 averaged over 3 runs. The task model is DeepLab-v2.}
    \label{tab:ssl_hyperparameters}
    \setlength{\tabcolsep}{1.5mm}{
    \begin{tabular}{lccccc}
      \toprule 
       & & \makecell[c]{Semantic} & \makecell[c]{Real Image} & \makecell[c]{Portrait Image} & \makecell[c]{Night Image}\\
       \textbf{\;Methods} & & \makecell[c]{Segmentation} & \makecell[c]{Denoising} & \makecell[c]{Matting} & \makecell[c]{Enhancement} \\
      \midrule
      \;MT \cite{MeanTeacher} & $\lambda_{MT}$  & $1.00$ & $1.00$ & $1.00$ & $1.00$ \\
       & $\eta_{MT}$ & $3$ & $5$ & $5$ & $5$ \\
       & $\alpha_{MT}$ & $0.99$ & $0.99$ & $0.99$ & $0.99$ \\
      \midrule
      \;S4L \cite{S4L} & $\lambda_{S4L}$ & $0.10$ & $1.00$ & $1.00$ & $1.00$ \\
      \midrule
      \;AdvSSL \cite{AdvSemiSeg} & $\lambda^{l}_{Adv}$ &  $0.01$ & $0.001$ & $0.01$ & $0.001$ \\
       &  $\lambda^{u}_{Adv}$ &  $0.001$ & $0.001$ & $0.01$ & $0.001$ \\
      \midrule
      \;GCT (Our)  & $\lambda_{fc}$
       & $1.00$ & $0.10$ & $1.00$ & $0.10$ \\
       & $\lambda_{dc}$ & $100$ & $1.00$ & $100$ & $1.00$ \\
       & $\eta_{dc}$ & $3$ & $5$ & $5$ & $5$ \\
       & $\xi$ & $0.60$ & $0.60$ & $0.40$ & $0.60$ \\

        % & \multicolumn{5}{c}{Results (mIOU) of DeepLab-v2} \\
    %   \midrule

      \bottomrule
    \end{tabular}}
  \end{center}
  \vspace{-0.5cm}
\end{table}

\section*{Appendix D: Visual Comparisons}
Here we provide visual comparisons of the SSL results for four validated tasks. The red bounding box in the figure highlights some main differences in the outputs. As shown below, GCT surpasses existing SSL methods in visual effects.

\vspace{-0.5cm}
\begin{figure}[H]
\begin{center}
   \includegraphics[width=0.90\linewidth]{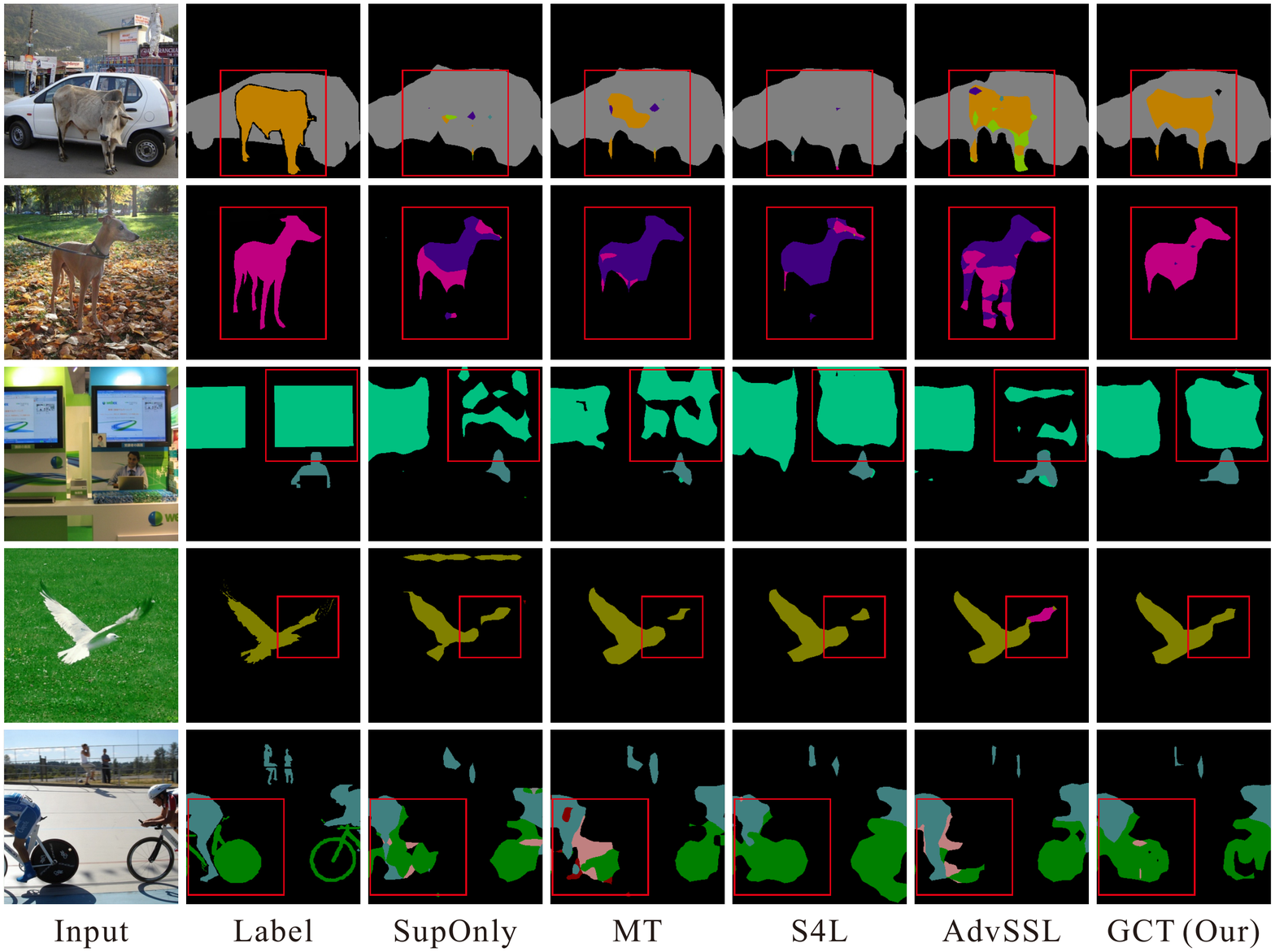}
\end{center}
\vspace{-0.7cm}
   \caption{\textbf{Semantic Segmentation.} Comparisons on the PASCAL VOC dataset using $1/8$ labeled data. 
    }
\label{fig:flaw_map}
\end{figure}

\vspace{-0.5cm}
\begin{figure}[H]
\begin{center}
   \includegraphics[width=0.90\linewidth]{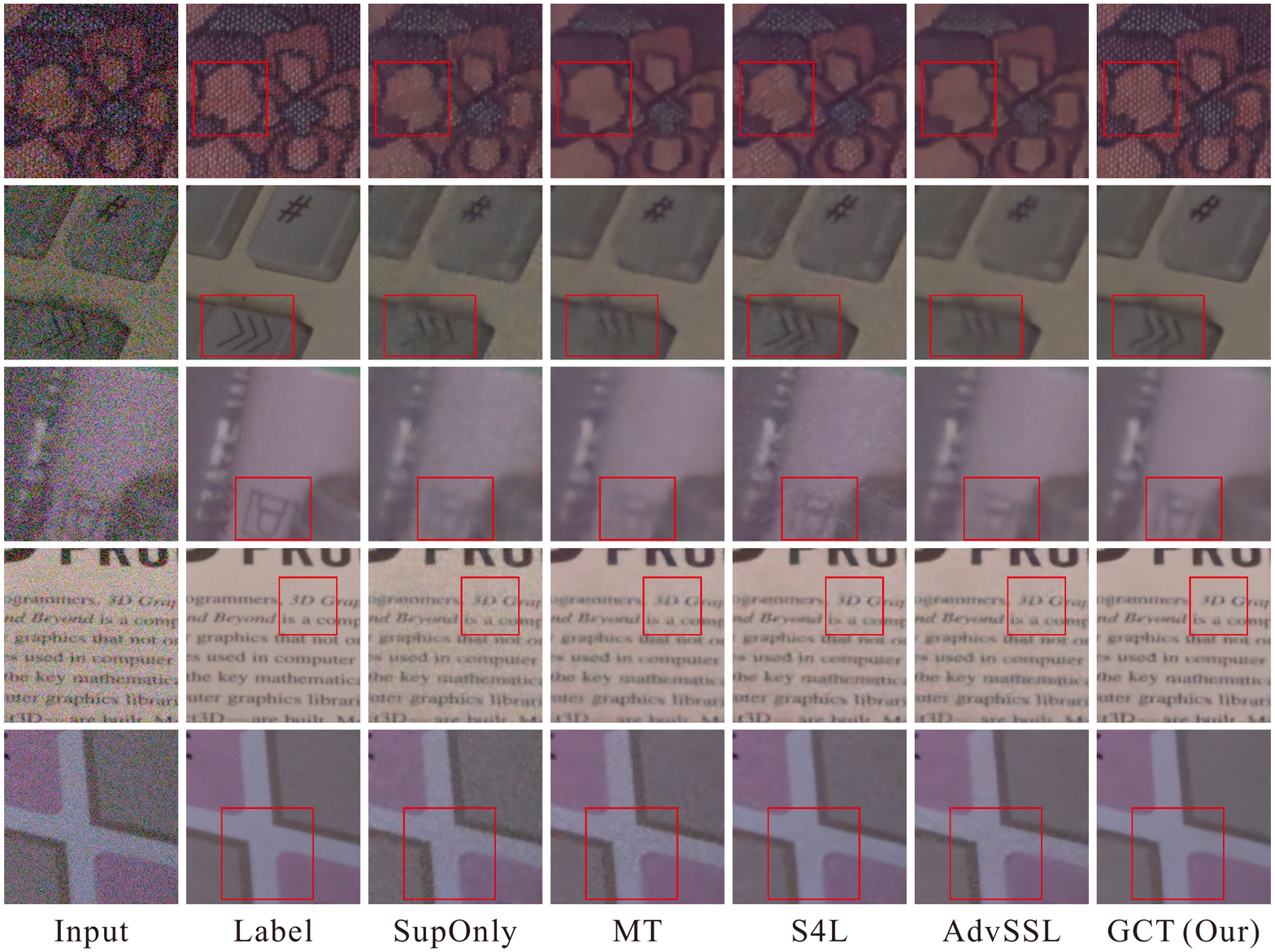}
\end{center}
\vspace{-0.7cm}
   \caption{\textbf{Real Image Denoising.} Comparisons on the SIDD dataset using $1/8$ labeled data. 
    }
\label{fig:flaw_map}
\end{figure}

\vspace{-1.5cm}
\begin{figure}[H]
\begin{center}
   \includegraphics[width=0.9\linewidth]{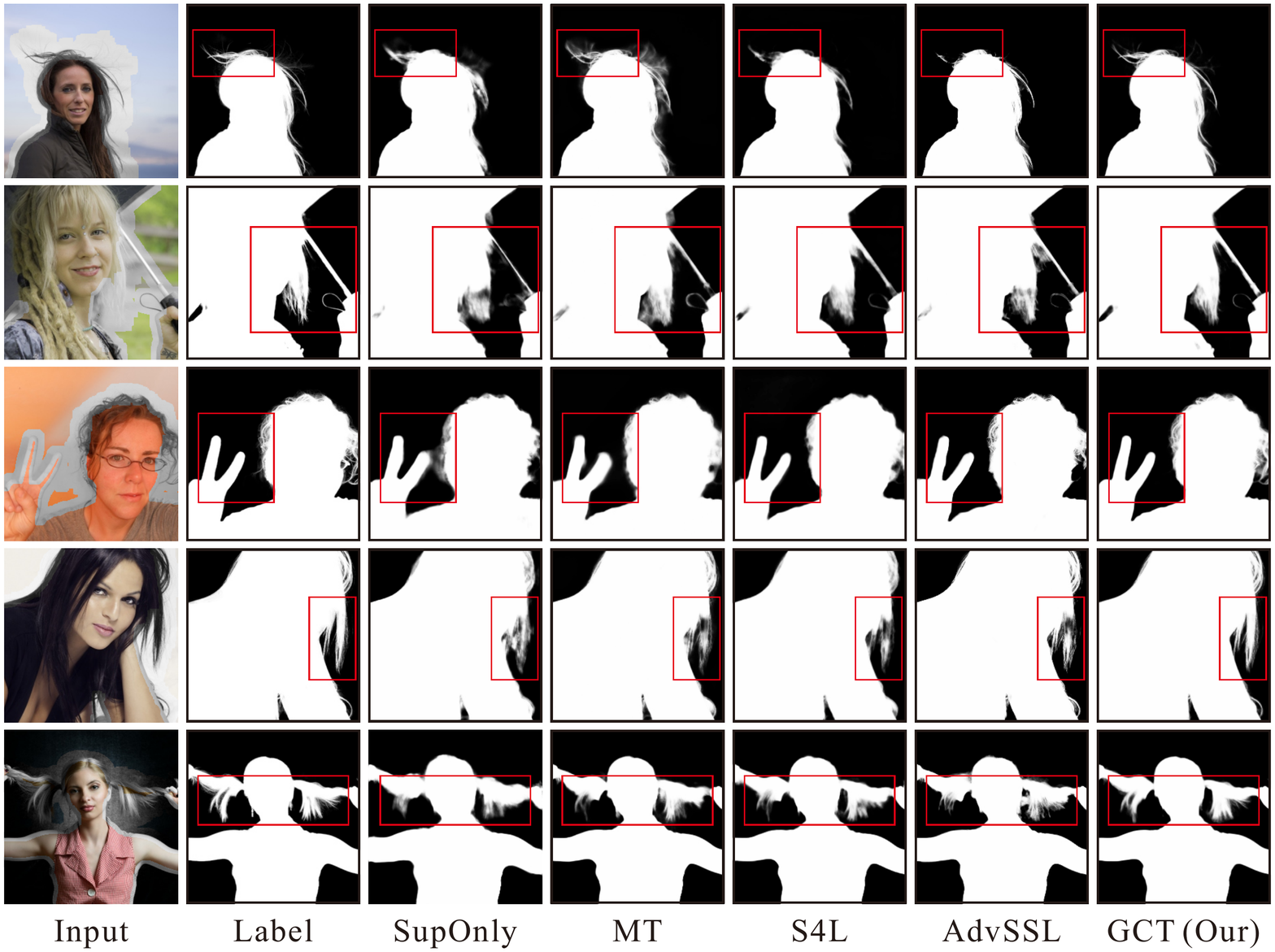}
\end{center}
\vspace{-0.7cm}
   \caption{\textbf{Portrait Image Matting.} Comparisons on our dataset using $100$ labeled data and $3850$ unlabeled data. 
    }
\label{fig:flaw_map}
\end{figure}

\begin{figure}[H]
\begin{center}
   \includegraphics[width=0.9\linewidth]{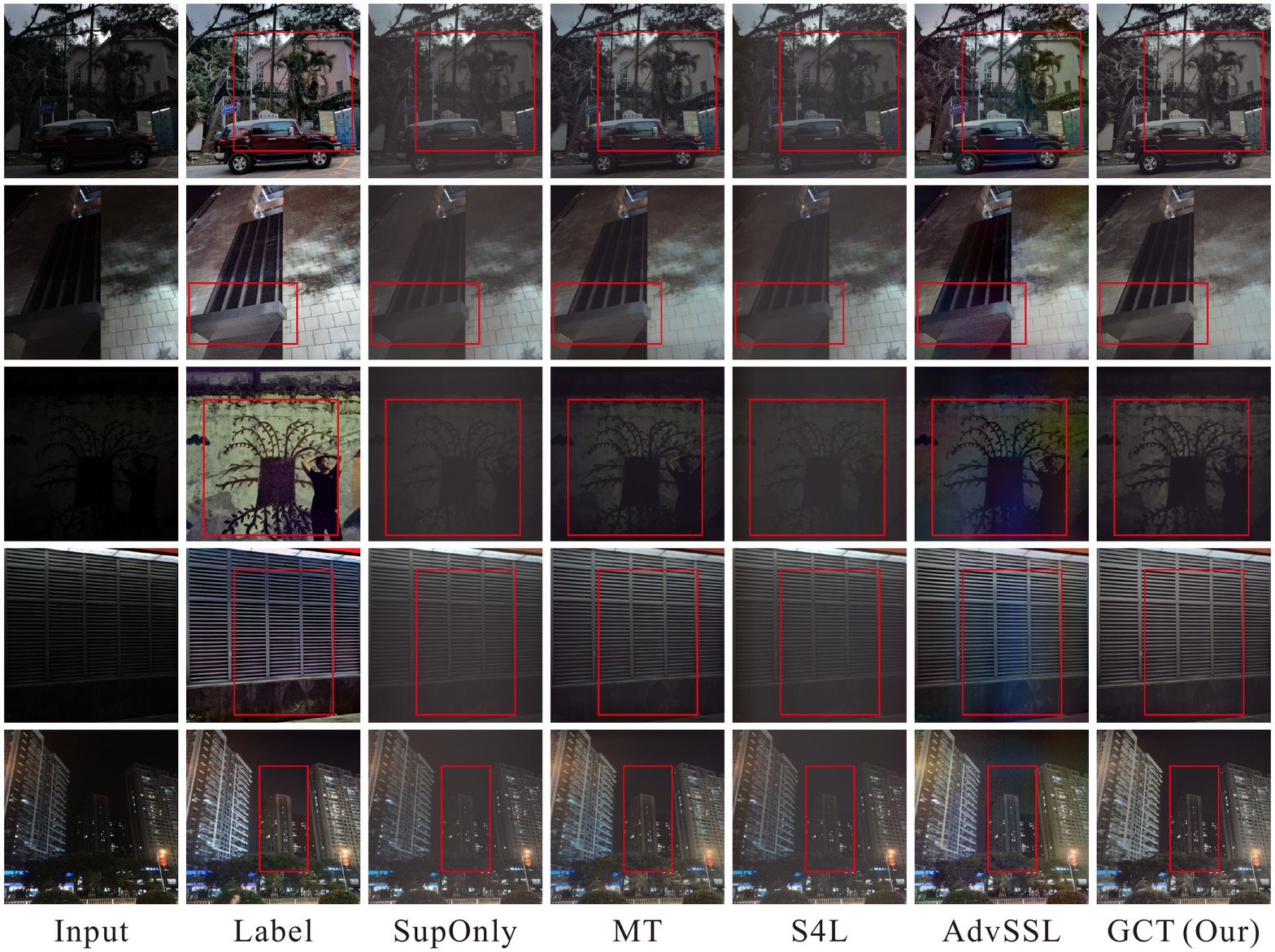}
\end{center}
\vspace{-0.7cm}
   \caption{\textbf{Night Image Enhancement.} Comparisons on our dataset using $200$ labeled data and $1500$ unlabeled data. 
    }
\label{fig:flaw_map}
\end{figure}

% \clearpage

% ---- Bibliography ----
%
% BibTeX users should specify bibliography style 'splncs04'.
% References will then be sorted and formatted in the correct style.
%
\bibliographystyle{splncs04}
\bibliography{egbib}
\end{document}